\pdfoutput=1
\documentclass[letterpaper, 10 pt, conference, left=1in,right=1in,top=1in,bottom=1.5in]{IEEEtran}  

\IEEEoverridecommandlockouts                              

\usepackage{graphics} 
\usepackage{epsfig} 
\usepackage{mathptmx} 
\usepackage{times} 
\usepackage{amsmath} 
\usepackage{amssymb}  

\usepackage[T1]{fontenc}
\usepackage{amsfonts}
\usepackage{booktabs}
\usepackage{siunitx}
\usepackage{caption} 
\usepackage{booktabs}
\usepackage{graphicx}
\usepackage{algorithmic}
\usepackage[ruled]{algorithm2e}
\usepackage{float}
\usepackage[percent]{overpic}

\usepackage{mwe,tikz}\usepackage[percent]{overpic}

\usepackage{subcaption}
\usepackage{array,   tabularx, makecell, booktabs}
\usepackage{geometry}
\usepackage{comment}

\title{\LARGE \bf
All-in-One: A DRL-based Control Switch Combining State-of-the-art Navigation Planners
}

\author{Linh K{\"a}stner$^{1}$\thanks{$^{1}$Linh K{\"a}stner, Johannes Cox, Teham Buiyan, and Jens Lambrecht are with the Chair Industry Grade Networks and Clouds, Faculty of Electrical Engineering, and Computer Science,				
		Berlin Institute of Technology, Berlin, Germany
		{\tt\small linhdoan@tu-berlin.de}}, Johannes Cox$^{1}$, Teham Buiyan$^{1}$, and Jens Lambrecht$^{1}$
}

\begin{document}

\maketitle
\thispagestyle{empty}
\pagestyle{empty}


\begin{abstract}

Autonomous navigation of mobile robots is an essential aspect in use cases such as delivery, assistance or logistics. Although traditional planning methods are well integrated into existing navigation systems, they struggle in highly dynamic environments. On the other hand, Deep-Reinforcement-Learning-based methods show superior performance in dynamic obstacle avoidance but are not suitable for long-range navigation and struggle with local minima.
In this paper, we propose a Deep-Reinforcement-Learning-based control switch, which has the ability to select between different planning paradigms based solely on sensor data observations.
Therefore, we develop an interface to efficiently operate multiple model-based, as well as learning-based local planners and integrate a variety of state-of-the-art planners to be selected by the control switch. Subsequently, we evaluate our approach against each planner individually and found improvements in navigation performance especially for highly dynamic scenarios. Our planner was able to prefer learning-based approaches in situations with a high number of obstacles while relying on the traditional model-based planners in long corridors or empty spaces.

\end{abstract}


\section{Introduction}
With the rise of mobile robotics for tasks such as logistics, health care, or last-mile delivery, navigation within unknown and highly dynamic environments has become an essential aspect \cite{fragapane2020increasing}, \cite{alatise2020review}.
Current industrial navigation solutions often employ hierarchical planning systems consisting of a global planner for long-range planning in static environments, and a local planner considering unknown and dynamic obstacles. However, these approaches reach their limits when dealing with fast-moving obstacles and require large computational efforts.
Typically, to ensure safety and robustness, conservative safety thresholds, hand-engineered safety measures, or the complete avoidance of dynamic zones are being employed, which makes the operation of mobile robots inflexible and limited.
Deep Reinforcement Learning (DRL) has emerged as an end-to-end planning method to replace overly conservative approaches in highly dynamic environments due to its ability to achieve efficient navigation and obstacle avoidance from raw sensor input. Thus, a variety of researchers employed DRL for obstacles avoidance \cite{dugas2020navrep}, \cite{guldenringlearning}, \cite{everett2018motion}, \cite{chen2019crowd}.
However, DRL-based approaches are prone to local minima, which makes specific situations such as long corridors, corners, or dead-ends challenging and sometimes unsolvable.
Efforts to increase the capabilities of DRL-based systems to achieve long-range navigation are complex and intensify the already tedious training procedures \cite{faust2018prm}, \cite{chiang2019learning}.
A hybrid approach to leverage the strengths of multiple planners would increase flexibility, safety, and efficiency.

\begin{figure}[]
    \centering
    \includegraphics[width=0.45\textwidth]{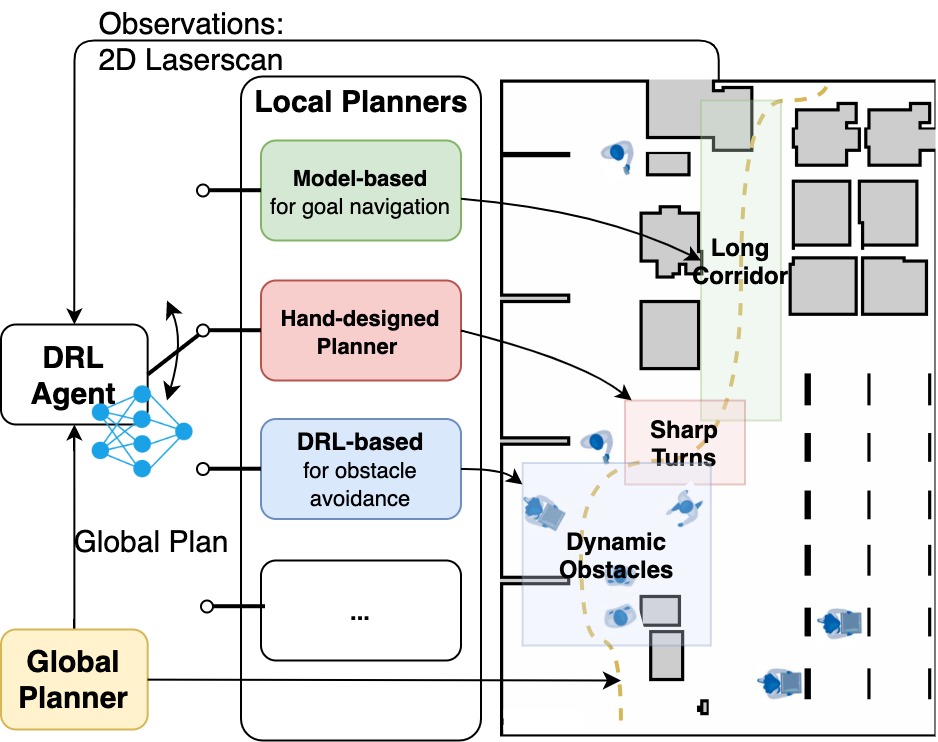}
    \caption{Our proposed all-in-one planner combines a variety of state-of-the-art planning approaches and trains an agent to decide which approach should be inferred in each specific situation. For instance, a DRL-based planner is inferred when there are multiple dynamic obstacles approaching, while a classic model-based planner is inferred in long corridors or maze-office environments. This way, the advantages of multiple planning approaches can be leveraged.}
    \label{intro}
\end{figure}
\noindent On that account, we combine both model- and learning-based approaches by proposing a DRL agent, which is trained to decide and switch between multiple different local planners based solely on sensor data observations. We name the control switch All-in-One agent (AIO) and train it on our efficient 2D simulator arena-rosnav of our previous work \cite{kastner2021towards}. The agent should be trained to decide the most suitable planner for every specific situation. Subsequently, we benchmark and compare our proposed AIO planner against each of its components individually in terms of safety, efficiency, and robustness.
The main contributions of this work are the following:
\begin{itemize}
    \item Integration of both model- and learning-based local planners into a hierarchical navigation system
    \item Proposal of a DRL-based control switch to decide between multiple local planners for each situation based solely on 2D laser scan observations
    \item Extensive evaluation of the control switch against benchmark navigation systems in terms of safety, efficiency, and robustness

\end{itemize}
The paper is structured as follows. Sec. II begins with related works. Subsequently, the methodology is presented in Sec. III. Sec. IV presents the results and discussion. Finally, Sec. V will give a conclusion and outlook. We made the code publicly available under https://github.com/ignc-research/all-in-one-DRL-planner.

\section{Related Works}
Autonomous navigation of mobile robots has been extensively studied in various research publications. Whereas navigation in static environments can be solved by using traditional approaches such as A-Star or RRT, dynamic obstacle avoidance is still an open frontier.
Traditional methods regard the obstacle avoidance task as an optimization control problem \cite{rosmann2015timed}, \cite{rosmann2019time}, \cite{fox1997dynamic}. These approaches require high computational calculations and can not cope well with fast-moving obstacles. On the other hand, learning-based approaches such as DRL-based navigation demonstrated superior performance for obstacle avoidance \cite{chen2017decentralized}. \cite{chen2019crowd}, \cite{dugas2020navrep}, \cite{guldenringlearning}, \cite{faust2018prm} but struggle in situations like mazes, corners, dead-ends or long corridors. Thus a variety of research works propose hybrid approaches, which aim to leverage the strength of multiple planners. The concept of switching between an array of different controllers is widely known and used in control theory. Often, these control problems propose a switching mechanism using, for instance, probability distributions \cite{milutinovic2018markov}, Lyapunov functions \cite{toibero2011switching}, or hand-designed rules \cite{jin2017stable}.
Jin et al. \cite{jin2017stable} proposed a switching approach for wheeled mobile robots combining goal-navigation and obstacle avoidance by defining an avoidance vector and hand-designed switching rules. Despite the improved performance in safety, the approach is not flexible and needs to be hand-designed for each scenario.
Similarly, Malone et al. \cite{malone2017hybrid} proposed a switching controller using stochastic reachable sets.
A large number of these approaches were proposed by control theory and include complex underlying theoretical assumptions. Furthermore, they are not tested on actual robots and only validated on specific use cases such as simple velocity control of a bicycle or parking situations.
In works from Jacobson et al.  \cite{jacobson2021approximate} the researchers proposed a switching between different behaviors using Filippov conditions and stochastic potential fields to optimize the gap between switching operations. The researcher could successfully deploy their controller towards a Turtlebot 3 and validate the switching approach in simulation.
In the works from Berkane et al. \cite{berkane2019hybrid}, a hybrid controller is proposed that is able to switch between a robust motion-to-goal and obstacle avoidance mode using traditional approaches. The main drawback is that the researchers require a variety of presumptions and have only tested the approach in simple simulation environments.
Most similar to our work, Cimurs et al. \cite{cimurs2020goal} and Zhang et al. \cite{zhang2019behavior} propose a DRL-based controller to switch between different modes. However, the approach was only demonstrated on simple, static scenarios without moving obstacles.
Instead of combining low-level controllers, our work focuses on the usage of high-level planners and train a DRL agent in an end-to-end manner. The agent should learn to decide on the most suitable planner in each situation based solely on sensor observations. This makes our approach more flexible against incoming rapid changes and enables more efficient and safe navigation in highly dynamic environments as we leverage the strength of DRL for dynamic obstacle avoidance as well as model-based approaches for long-range navigation.

\begin{figure*}[]
	\centering
	\includegraphics[width=0.96\textwidth]{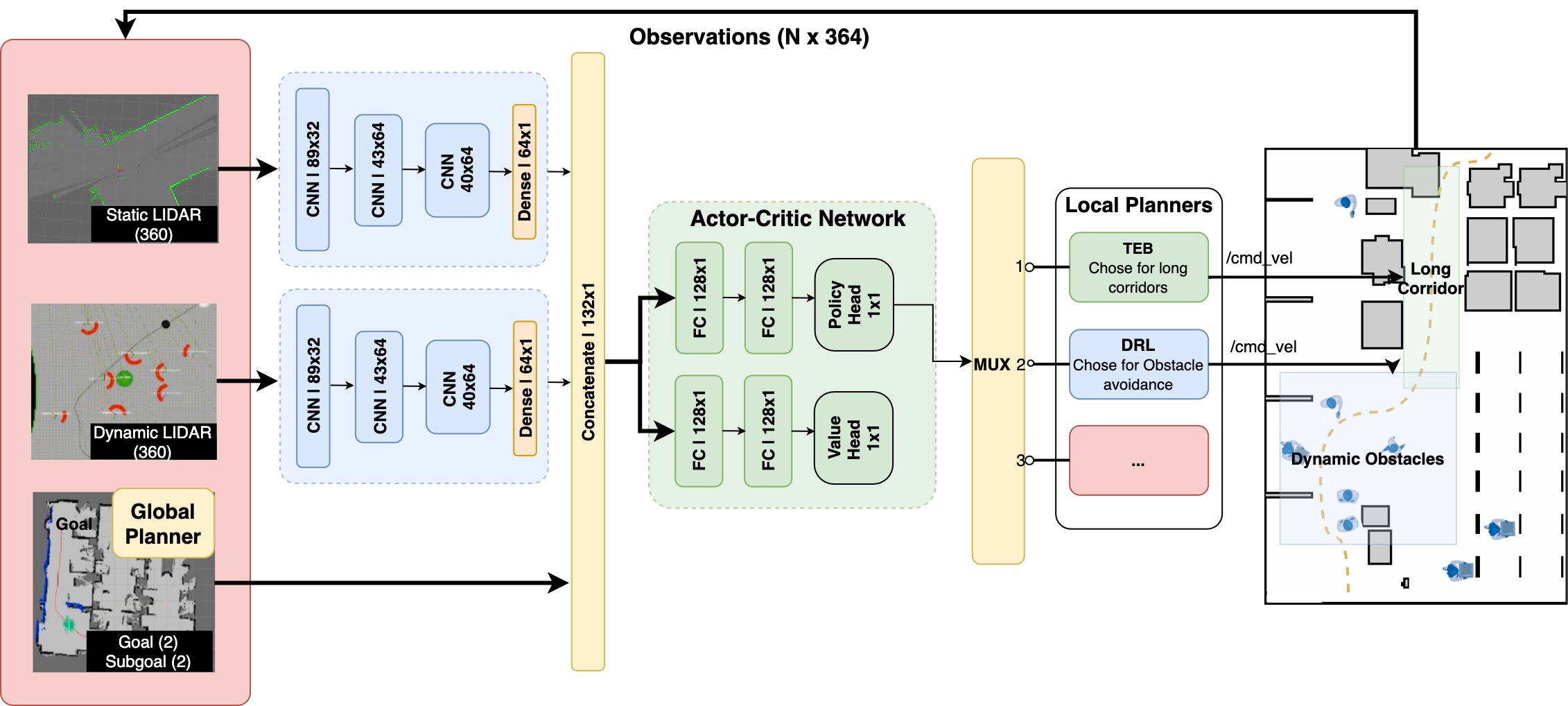}
	\caption{System design of our proposed approach. The observation space of our agent consists of two types of laser scan representations to indicate dynamic and static observations and the goal. The input is processed separately before concatenation. We employ an actor-critic approach where the agent outputs a discrete value selecting from an array of possible local planners, which navigates to the goal. An interface is implemented to only infer the respective planners when the agent provides a trigger signal. Finally, the agent learns to select the most suitable planners for each situation.}
	\label{system}
\end{figure*}

\section{Methodology}
In this chapter, we present the methodology of our proposed framework. Our system design is illustrated in Fig. \ref{system}. We train a DRL agent to choose between multiple different local planners and select the most suitable one for any specific situation based solely on laser scan observations.

\subsection{System Design}
To enable the fast switching between the planners, we implemented an interface, which allows to trigger a planner instantly and compute the action commands.
The interface is based on the move-base package,  the difference being that the planners can be called via services instead of running periodically.
Therefore, the necessity to run all planners in parallel is eliminated, which enables the system to only activate the planner if it is selected by the DRL agent. This makes the system more efficient.
To combine the advantages of model-based planners, namely navigation over long ranges, corridors, corners, etc. with that of DRL-based approaches namely obstacle avoidance, our interface is able to integrate a variety of both planning paradigms. In total, the interface provides the planners MPC \cite{rosmann2019time}, TEB \cite{rosmann2015timed}, DWA \cite{fox1997dynamic} as model-based planners, and CADRL \cite{everett2018motion}, RLCA \cite{chen2019crowd} and DRL \cite{kastner2021towards} of our previous work as learning-based planners. The interface can be extended to include more planners. The final selection of these planners will be described in the agent design section. The velocity bounds are set to $[-0.15,0.3]\frac{m}{s}$ for the linear velocity and $[-2.7,2,7]\frac{rad}{s}$ for the angular velocity. The minimum distance to all obstacles is set to 0.3m plus 0.2m inflation because the robot has to be able to navigate through narrow corridors. As a global planner, we employ the Dijkstra approach, which is also called using services. The replanning is running at $\frac{1}{10}$ times the frequency of the local planner (1s vs 100ms).

\subsection{Agent Design}
The observation space of our DRL agent contains the robot pose (through localization), the robot scan (360-degree laser scan sensor), the goal pose, the sub-goal pose (through sampling from the global path), and a known / static map (it is assumed that mapping was already performed). Subsequently, a static and dynamic scan representation is calculated. The static scan is calculated from the robot pose and the static map, whereas the dynamic scan $s_\text{dyn}$ is calculated from the static scan $s_\text{static}$ and the sensor scan $s_\text{sensor}$. Equation 1 outlines these calculations.
\begin{equation*}
    s_\text{dyn}(x)=
    \left\{
        \begin{array}{ll}
        s_\text{sensor}(x)  & \mbox{if } |s_\text{sensor}(x) - s_\text{static}(x)| > \epsilon \\
            \text{laser-range} + 1 & \mbox{otherwise }
        \end{array}
    \right.
\end{equation*}

\noindent Two different scan streams are used to provide the switch with information about the scenario in which the robot is in, such as 1) driving through an empty area 2) driving next to static obstacles, e.g. a corridor, or 3) driving next to dynamic obstacles. A distinction between scenarios 2) and 3) is not possible using only one laser scan measurement at a time. A pre-condition is that knowledge of the robot pose (localization) and the static map (mapping) can be extracted without large errors or uncertainties. It is noted that the dynamic and static scan is only used for the DRL agent switch and not for the local planners themselves. Therefore, in the case of an error in localization, the local planners still work as intended. The AIO agent selects a local planner from a given list of planners making the action space discrete and of the size of the list of local planners. The switch runs with a frequency of 2 Hz.

\subsubsection{Planner Selection}
After extensive testing and careful assessment of all planners, our AIO agent employs TEB as a model-based planner and a DRL-based approach from our previous work \cite{kastner2020deep} as a learning-based planner for obstacle avoidance. This is based on the following assessments: Rlca \cite{chen2017decentralized} is not competitive on office maps due to its hard-coded safe policy where the robot drives slowly when close to obstacles, which works well for multi-robot scenarios but does not work for static obstacles and can lead to collisions with dynamic obstacles, which do not follow that same policy. The MPC planner is computationally demanding and less stable compared to TEB. Moreover, the non-linear controlling aspect is not relevant for our differential drive robot model. The CADRL approach \cite{everett2018motion}  showed robust obstacle avoidance in highly dynamic environments. However, it was also designed for multi-robot systems and requires knowledge of all other obstacles. The DWA planner \cite{fox1997dynamic} performs worse than TEB with no advantages. \newline

\subsection{Neural Network Design}
Fig. \ref{system} illustrates the neural network design to process the observations and output recommendations to chose a planner. The network consists of a feature extractor for both scans separately based on a convolutional layer and one fully connected layer. The features are concatenated and given as input into two MLPs for the policy and the value function. Relu is used for the whole network.

\subsection{Reward System}
Our reward system is formalized in Eq. \ref{reward}. It gives a small reward for following the global plan $r_{\text{global\_plan}}$, a small reward for approaching the global goal $r_{\text{approaching\_goal}}$ and a medium negative reward when the robot is too close to an obstacle $r_{\text{safe\_dist}}$. This results in the following reward
\begin{align}
        r_{total} &= r_{\text{collision}} + r_{\text{goal\_reached}} + r_{\text{global\_plan}} \\
        &+ r_{\text{approaching\_goal}} + r_{\text{safe\_dist}}
        \label{reward}
\end{align}

In general, tuning the reward for collision and following the global plan will influence the selection choice of the AIO agent due to the trade-off between safety and efficiency. Since the main emphasis is robust collision avoidance, the reward for following the global path is moderate. Approaching a goal results in a reward based on the difference of the distance to the sub-goal between two time steps $\delta_\text{sub-goal}$:
\begin{equation}
    r_\text{approaching\_goal} =
    \left\{
        \begin{array}{ll}
            \delta_\text{sub-goal} \cdot 0.25  & \mbox{if } \delta_\text{sub-goal} >= 0 \\
            - \delta_\text{sub-goal} \cdot 0.5 & \mbox{if } \delta_\text{sub-goal} < 0
        \end{array}
        \right.
\end{equation}
The punishment is twice as high as the punishment for driving in circles.
\begin{equation}
    r_\text{safe\_dist} =
    \left\{
        \begin{array}{ll}
            -0.1  & \mbox{if } \min(s_\text{dyn}) < 0.18 \text{m} \\
            0  & \mbox{otherwise}
        \end{array}
        \right.
\end{equation}
The final reward is based on the possible outcomes: goal reached, collision occurred, or maximum number of iterations exceeded (mostly when the robot got stuck).
\begin{equation}
    r_\text{final} =
    \left\{
        \begin{array}{ll}
            3  & \mbox{if } \text{goal reached} \\
            0  & \mbox{if } \text{max iterations reached} \\
            -2 & \mbox{if } \text{collision occured}
        \end{array}
        \right.
\end{equation}

\subsection{Training procedure and environments}
\begin{figure}[!h]
	\centering
	\includegraphics[width=0.45\textwidth]{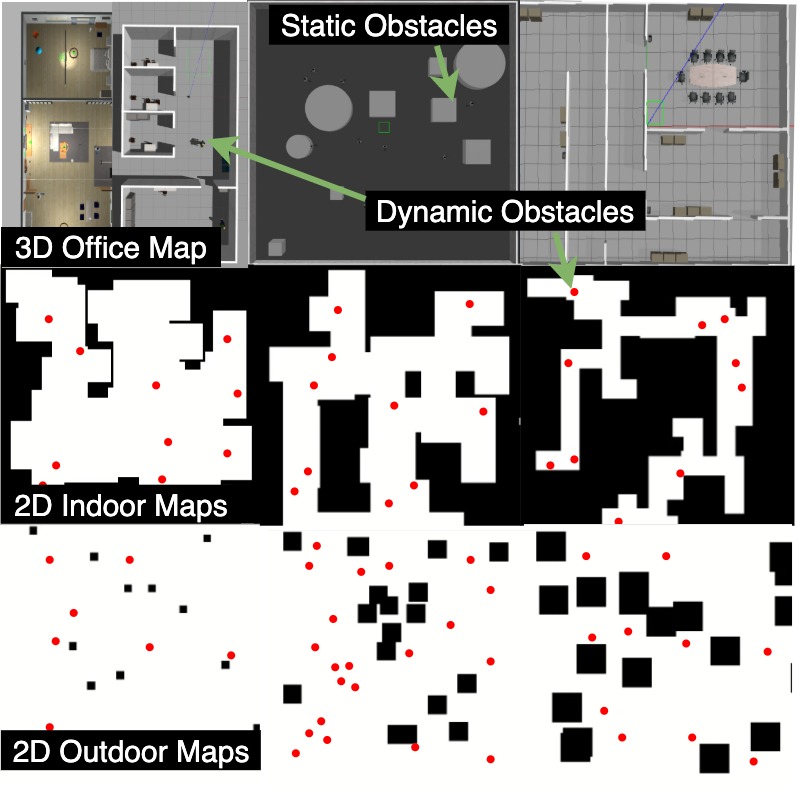}
	\caption{Test environments. We trained the agent on indoor and outdoor maps. The maps are randomly generated and increase in difficulty if a specific success threshold is reached. In indoor maps, the corridors become smaller while in outdoor maps, the static obstacles become larger. Both random, as well as static obstacles, are randomly spawned for each new episode.}
	\label{envs}
\end{figure}

The agent is trained on randomized maps to mitigate overfitting. The training environments are depicted in Fig. \ref{envs}. We employ a curriculum training, which increases in difficulty once a success threshold is reached. In indoor maps, the corridors become smaller while in outdoor maps, the static obstacles become larger. Both random, as well as static obstacles, are randomly spawned for each new episode. The experimental tests were done in both the 2D simulator with randomized maps containing static and dynamic obstacles.

\begin{figure*}[!h]
    
        \begin{subfigure}{0.33\textwidth}(a)
         \centering 
        \includegraphics[width=\linewidth]{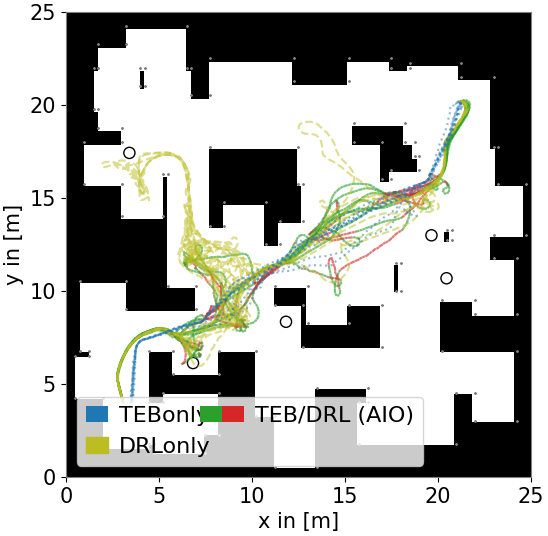}
        
        \label{fig:1}
    \end{subfigure}\hfil 
    \begin{subfigure}{0.33\textwidth }(b)
     \centering 
        \includegraphics[width=\linewidth]{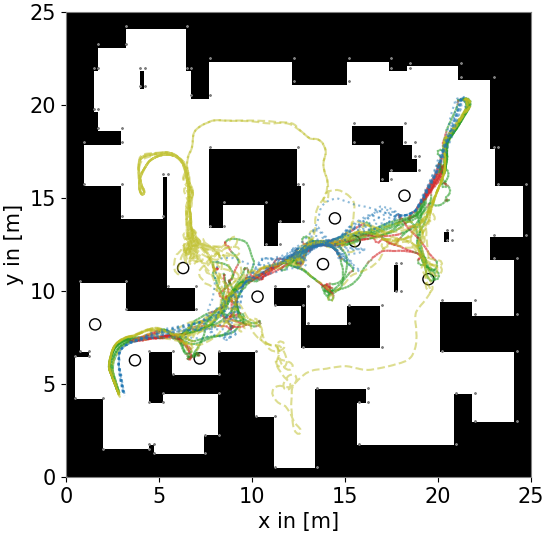}
        
        \label{fig:2}
    \end{subfigure}\hfil 
    \begin{subfigure}{0.33\textwidth}(c)
     \centering 
        \includegraphics[width=\linewidth]{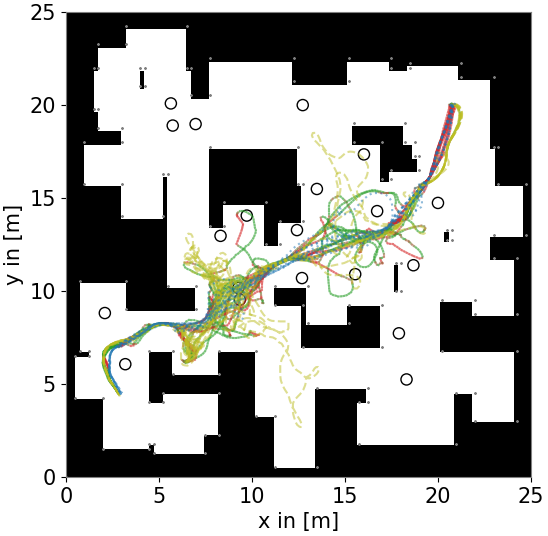}
        
        \label{fig:3}
    \end{subfigure}
    \medskip
    \begin{subfigure}{0.33\textwidth}(d)
         \centering 
        \includegraphics[width=\linewidth]{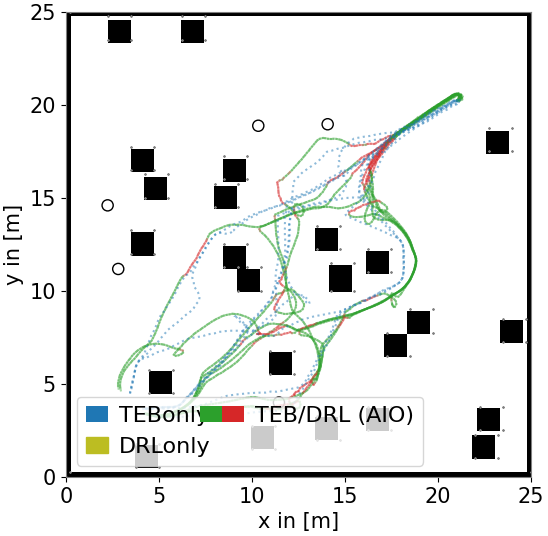}
        
        \label{fig:1}
    \end{subfigure}\hfil 
    \begin{subfigure}{0.33\textwidth}{e}
     \centering 
        \includegraphics[width=\linewidth]{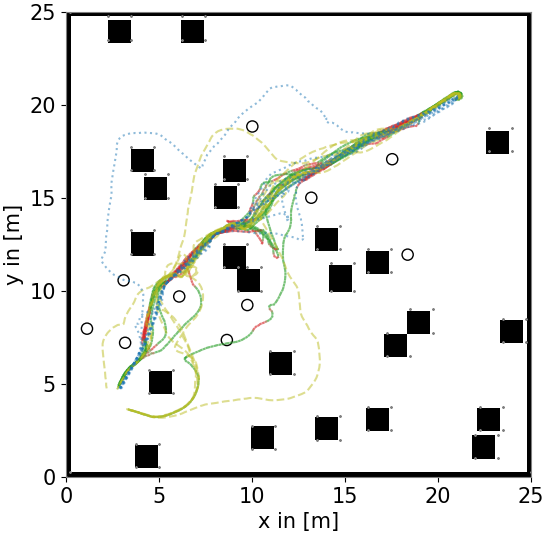}
        
        \label{fig:2}
    \end{subfigure}\hfil 
    \begin{subfigure}{0.33\textwidth}{f}
     \centering 
        \includegraphics[width=\linewidth]{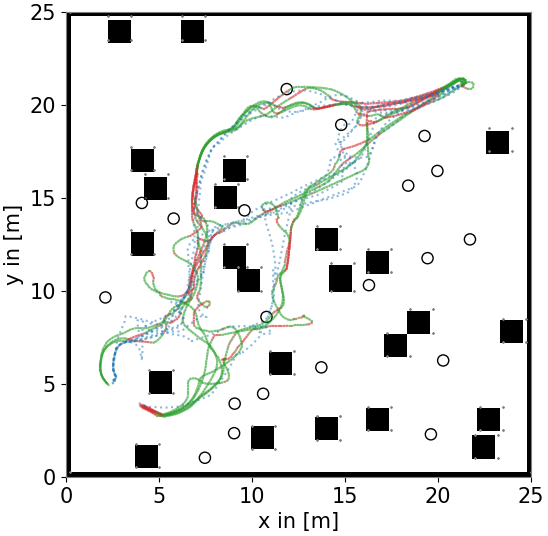}
        
        \label{fig:3}
    \end{subfigure}
    
    \caption{Trajectories of the AIO agent (red and green) and its standalone components: TEB (blue) and DRL (yellow) in the indoor map (upper row) and the outdoor map (lower row). The scenarios contain 5, 10, and 20 obstacles each with an obstacle velocity of 0.3 m/s.}
    \label{eval-quali}
\end{figure*}

\section{Evaluation}
In this chapter, we present the evaluation of our all-in-one agent denoted as AIO. We assess the navigation performance of our AIO agent both qualitatively and quantitatively against all our local planners individually. These include the model-based TEB, MPC, DWA, and RLCA, and DRL as learning-based baselines.

\subsection{Qualitative}
\begin{figure}[!h]
\begin{subfigure}{0.48\textwidth}(i) Indoor
         \centering 
        \includegraphics[width=\linewidth]{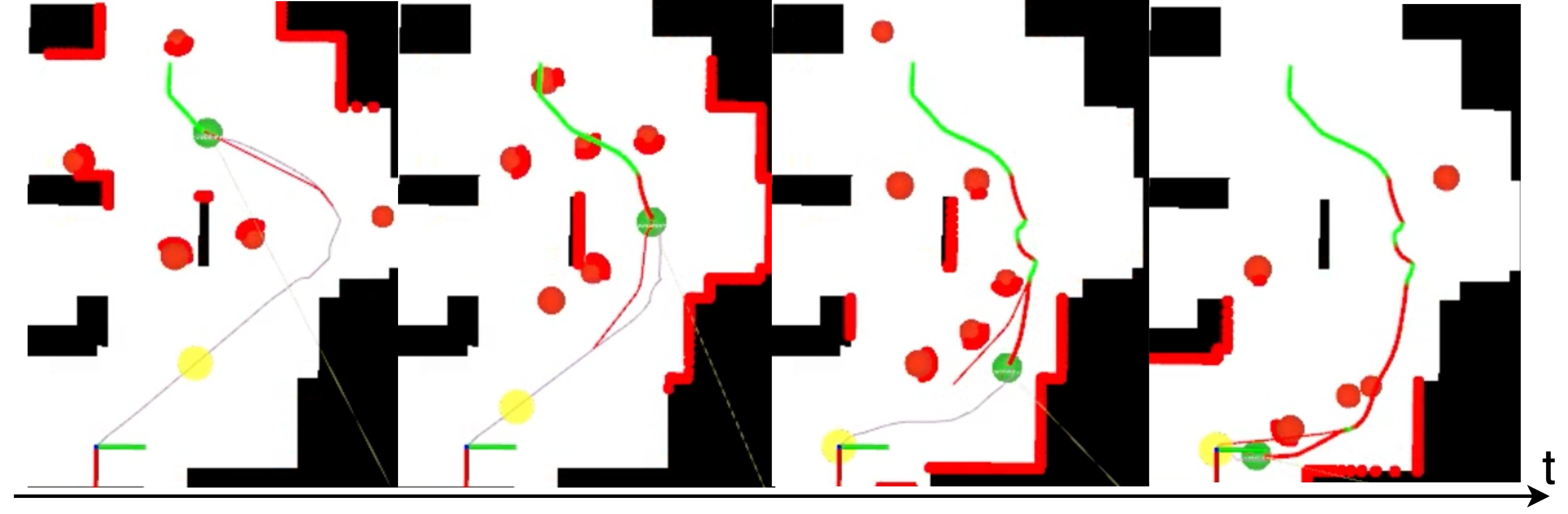}
        
        \label{fig:1}
    \end{subfigure}
    \medskip
    \begin{subfigure}{0.48\textwidth }(ii) Outdoor
     \centering 
        \includegraphics[width=\linewidth]{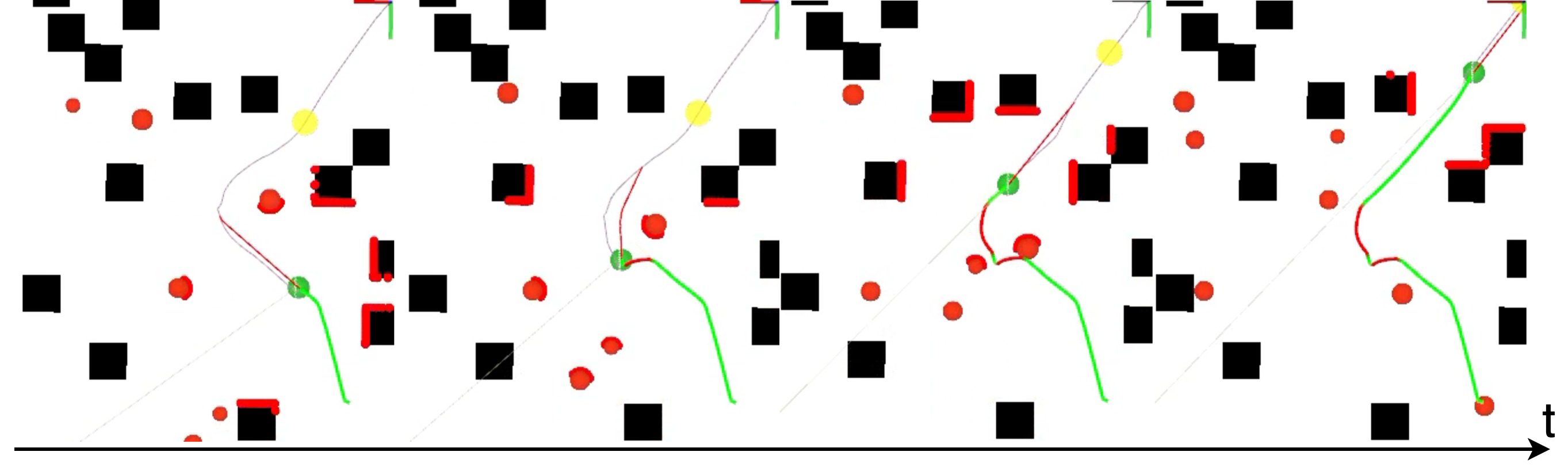}
        
        \label{fig:2}
    \end{subfigure}
    \caption{Trajectory of the AIO agent on the indoor and outdoor map over the time. The AIO agent switches to DRL (red) when encountering obstacles while utilizing TEB (green) in corridors and for situations with less or no obstacles. DRL is inferred more often in outdoor maps, whereas TEB is preferred in indoor maps due to a larger number of walls and corridors.}
    \label{office}
\end{figure}

For the qualitative evaluations, we deployed the AIO agent on indoor and outdoor scenarios with an increasing number of obstacles (5, 10, and 20 obstacles plotted as circles). Subsequently, we deployed the TEB and DRL approaches individually and recorded 20 runs for each. The results are illustrated in Fig. \ref{eval-quali}.
For our AIO agent, the trajectory is visualized with a specific color for TEB (green) and DRL (red). In the same plot, the DRL agent (yellow) and the TEB agent (blue) are plotted.
We can observe that our AIO planner is able to switch between model-based and learning-based approaches in specific situations. For instance, when there are multiple obstacles approaching, our AIO planner is able to select the learning-based approach, while in corridor situations, TEB is preferred. This leads to more straightforward and efficient trajectories towards the goal as well as fewer collisions in both indoor and outdoor environments. On the other side, the stand-alone DRL approach struggles in indoor environments, which results in a high number of roundabout paths as can be observed in Fig. \ref{eval-quali} (a) and (b). Contrarily, the stand-alone TEB approach accomplished a straightforward and robust path to the goal. In the outdoor scenarios, our AIO planner also selects between the two options and chooses the DRL planner when encountering obstacles e.g. in Fig. \ref{eval-quali} d) at position (x:17, y:16) or in e) at position (x:18,y:19), and TEB for goal-navigation. In Fig. \ref{office}, the trajectory of the AIO agent on both maps is plotted over time. Once again, the illustrations indicate the successful switching mechanism of our AIO agent, which chooses to switch to the DRL planner for incoming obstacles and otherwise relies on the TEB planner. The navigation behavior of our approach is demonstrated visually in the supplementary video.

\begin{figure*}[!h]
\begin{subfigure}{0.24\textwidth}(i)
         \centering 
        \includegraphics[width=\linewidth]{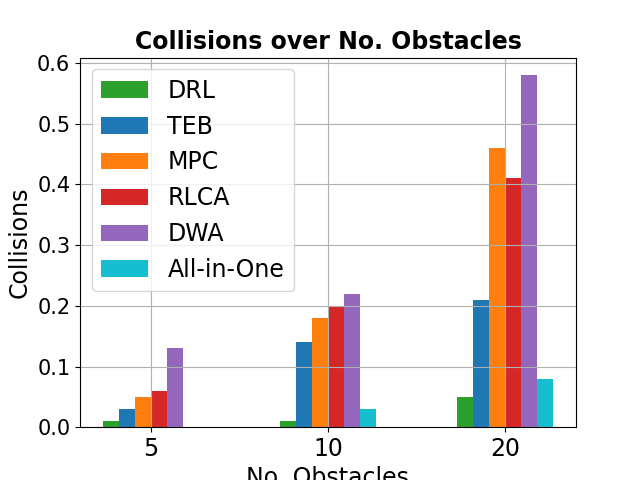}
        
        \label{fig:1}
    \end{subfigure}\hfil 
    \begin{subfigure}{0.24\textwidth }(ii)
     \centering 
        \includegraphics[width=\linewidth]{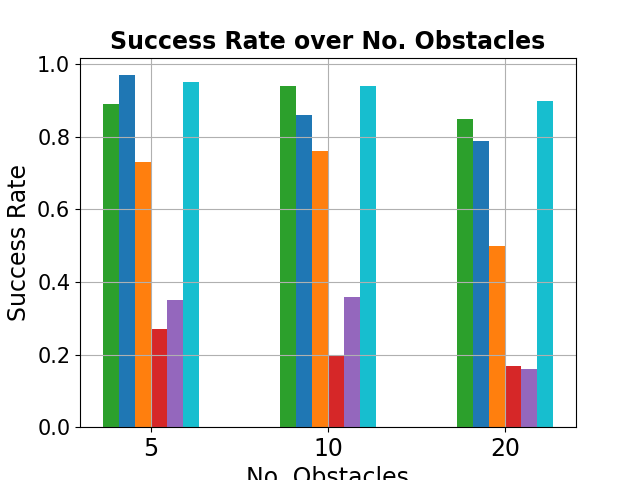}
        
        \label{fig:2}
    \end{subfigure}\hfil 
    \begin{subfigure}{0.24\textwidth}(iii)
     \centering 
        \includegraphics[width=\linewidth]{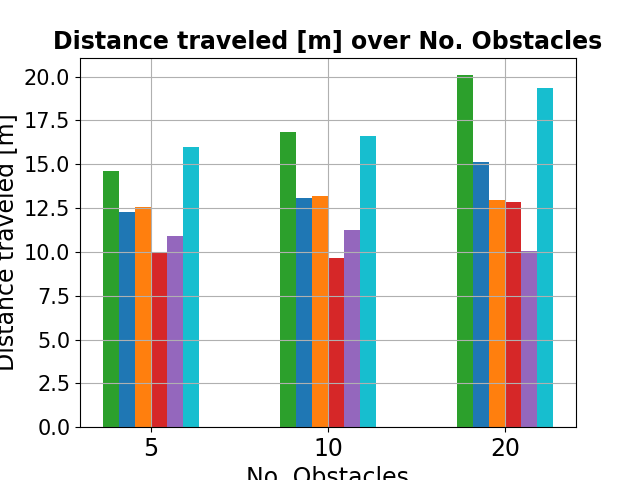}
        
        \label{fig:3}
    \end{subfigure}\hfil 
    \begin{subfigure}{0.24\textwidth}(iv)
     \centering 
        \includegraphics[width=\linewidth]{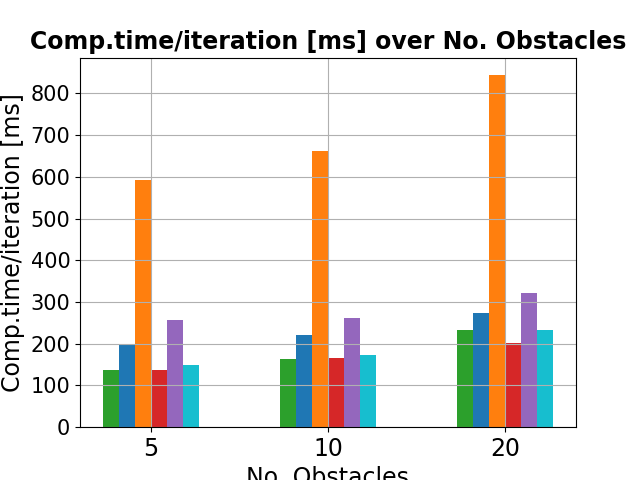}
        
        \label{fig:3}
    \end{subfigure}
    \medskip
    \begin{subfigure}{0.24\textwidth}(v)
         \centering 
        \includegraphics[width=\linewidth]{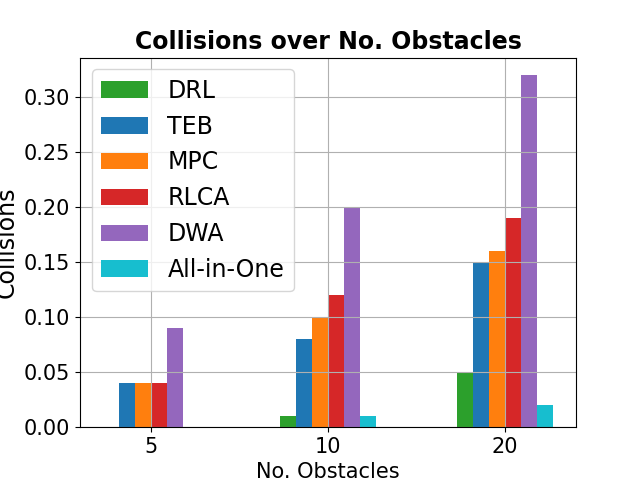}
        
        \label{fig:1}
    \end{subfigure}\hfil 
    \begin{subfigure}{0.24\textwidth }(vi)
     \centering 
        \includegraphics[width=\linewidth]{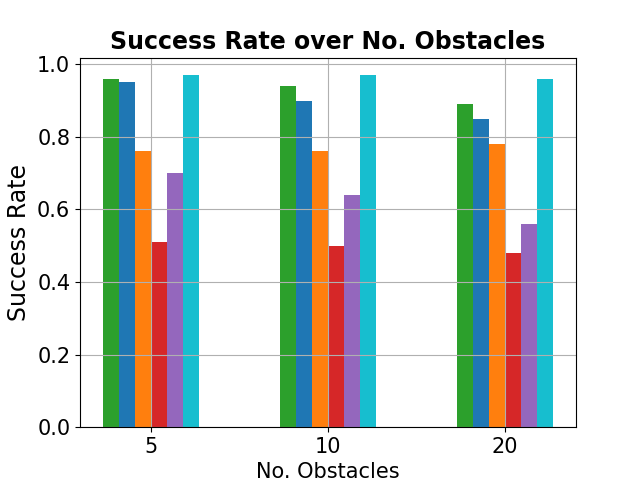}
        
        \label{fig:2}
    \end{subfigure}\hfil 
    \begin{subfigure}{0.24\textwidth}(vii)
     \centering 
        \includegraphics[width=\linewidth]{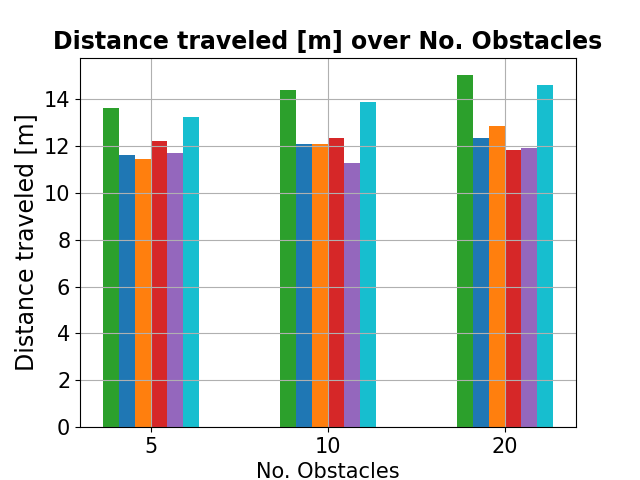}
        
        \label{fig:3}
    \end{subfigure}\hfil 
    \begin{subfigure}{0.24\textwidth}(viii)
     \centering 
        \includegraphics[width=\linewidth]{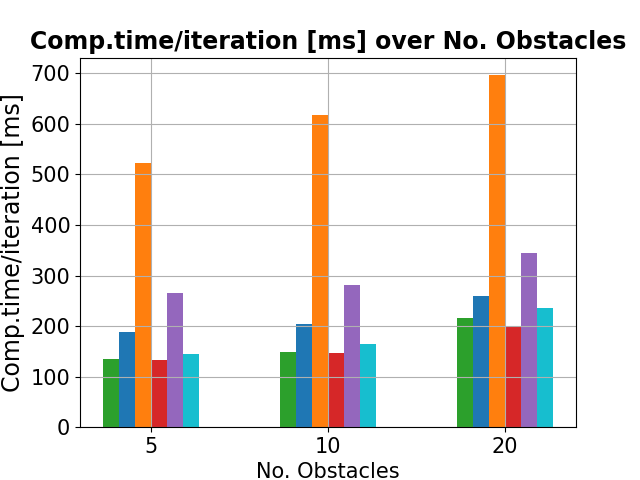}
        
        \label{fig:3}
    \end{subfigure}

    \caption{Quantitative evaluations of our AIO planner against all baseline approaches. The collisions, success rates, distances traveled, and computation times per iteration were plotted over the number of obstacles (5, 10, and 20) for indoor and outdoor scenarios. A total of 500 runs were conducted and the average values were calculated.}
    \label{quanti}
\end{figure*}

\subsection{Quantitative}
For the quantitative evaluation, we statistically tested our AIO agent and its individual planners TEB and DRL on randomly generated maps of increasing difficulty and obstacle numbers (5, 10, and 20 obstacles) over 500 episodes. Subsequently, we calculated the average success rates, the average collision rate, the average path length to reach the goal as well as the computation time per iteration. Additionally, we evaluated the baseline approaches MPC \cite{rosmann2019time}, RLCA \cite{chen2017decentralized}, and DWA \cite{fox1997dynamic}. The results are plotted in Fig. \ref{quanti}.
It is observed that our AIO planner outperforms all planners, especially in highly dynamic scenarios. 
Whereas the number of collisions increases for all planners, the AIO agent accomplishes a low collision rate with an average collision rate per episode of under 0.1 in the scenario with 20 obstacles. The model-based baselines MPC and DWA perform worst with over 0.45 and 0.55 average collisions per episode, respectively. In terms of robustness, the TEB planner performs best in scenarios with 5 obstacles accomplishing almost 99 percent, whereas our AIO planner reaches a 98 percent success rate. However, in scenarios with 20 obstacles, the success rate drops to 78 percent while our AIO planner still accomplishes 89 percent. The DRL agent also achieves competitive performance with a constant success rate of over 80 percent for all scenarios, which demonstrates the robustness of DRL for obstacle avoidance. However, in terms of efficiency, the stand-alone DRL approach requires the longest paths to reach the goal with over 20 meters in the indoor scenario with 20 obstacles. 
In terms of the computation time per iteration, which indicates the computational efficiency of a planner, it is visible that the MPC planner is the least efficient requiring triple the amount of time compared to all other approaches. This is due to the additional optimization processes done by the MPC planner, which is not relevant for our differential drive robot. Our AIO planner, along with the other learning-based approaches RLCA and DRL, requires the least computation time with only about 100ms per iteration making it slightly more efficient than the stand-alone TEB approach, which requires about 50 ms more per iteration. The results demonstrated that our AIO planner is more efficient and accomplishes a higher success rate in scenarios with a high number of obstacles compared to the stand-alone components TEB and DRL. In the outdoor environments, similar results were accomplished except for the DRL approach, which expectantly accomplished a higher success rate. 

\subsection{Switching Probabilities}
\begin{figure}[!h]
    \centering
    \includegraphics[width=0.5\textwidth]{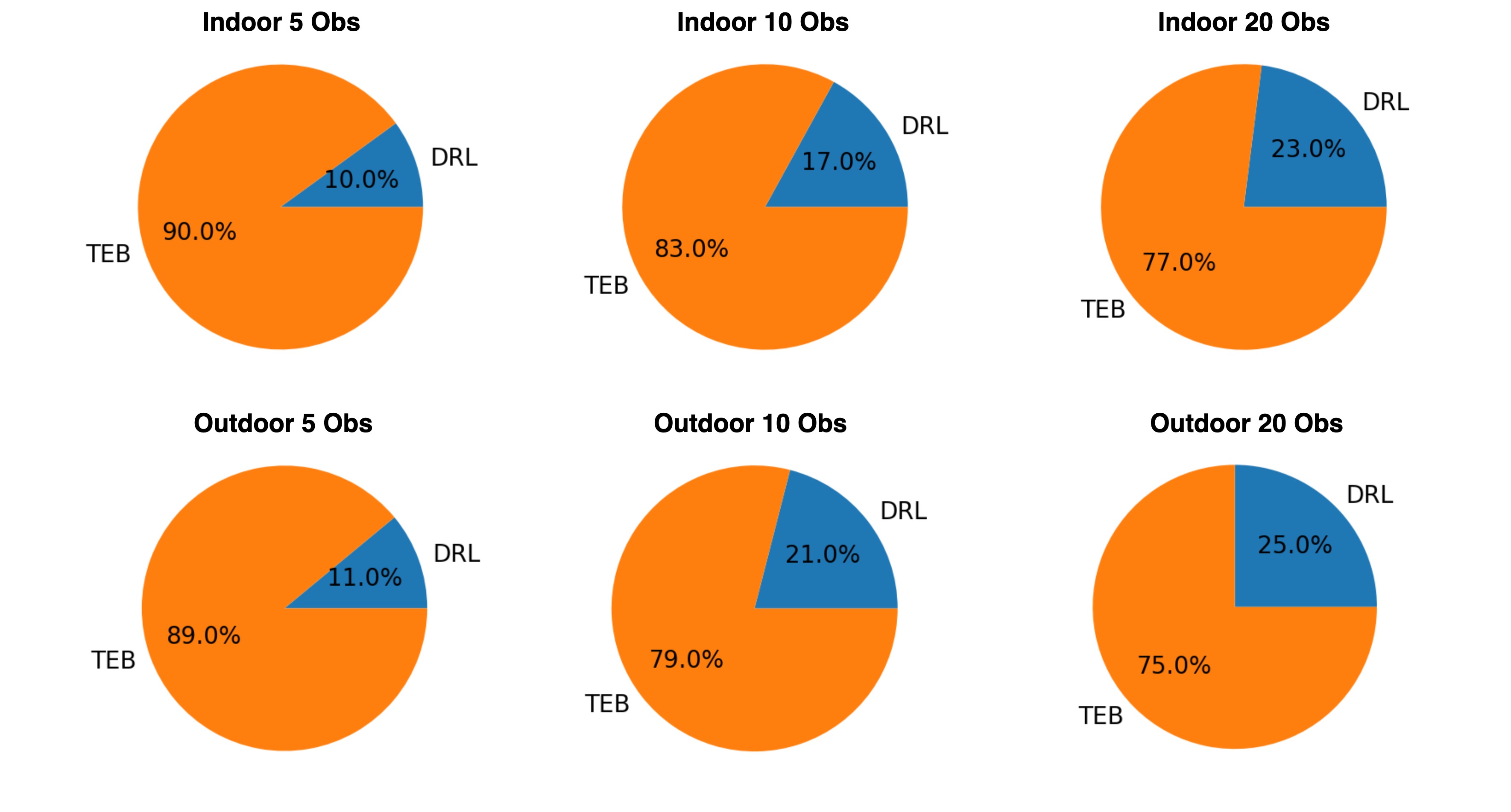}
    \caption{Percentages for each planner. While the DRL planner is infered more often in environments with a high number of obstacles and outdoors, the TEB planner is infered more often indoors and when there are no or few obstacles.}
    \label{piechart}
\end{figure}

Fig. \ref{piechart} illustrates the probabilities of each planner for the different scenarios. Our agent inferred the DRL planner slightly more often in outdoor environments. With an increasing number of obstacles, the usage of DRL increased to up to 25 percent in the outdoor scenarios with 20 obstacles. On the other hand, TEB was preferred by the agent in indoor environments as it coped better with long corridors and walls. TEB is also being used a majority of the time due to the more efficient planning utilizing the global and local costmap, which enables long-range navigation. Thus, the results indicate a clear preference for the DRL-based approach only for obstacle avoidance and the model-based TEB planner for navigation in static and indoor environments.

\section{Conclusion}
In this paper, we proposed a DRL-based control switch, which is able to select and switch between high-level planning approaches based on sensor observations to leverage the advantages of multiple methods. Therefore, we developed an efficient ROS navigation interface containing state-of-the-art model-based as well as learning-based approaches. We demonstrated that our AIO agent is able to select between different methods based on the specific situation. More specifically, DRL-based approaches were inferred in situations with a high number of obstacles while model-based approaches were selected for long-range navigation. Our approach was evaluated against the individual planners used in the AIO agent as well as other navigation systems. We found enhancement in safety, robustness, as well as efficiency using our AIO agent, especially in highly dynamic environments. Future works will extend the agent with a higher number of planners and the observation space with high-level semantic information. Moreover, we aspire to transfer and extensively evaluate our approach on real robots.

\appendix
The code is publicly available at https://github.com/ignc-research/all-in-one-DRL-planner



\addtolength{\textheight}{-1cm} 




\typeout{}
\bibliographystyle{IEEEtran}
\bibliography{ref}

\begin{thebibliography}{10}
\providecommand{\url}[1]{#1}
\csname url@samestyle\endcsname
\providecommand{\newblock}{\relax}
\providecommand{\bibinfo}[2]{#2}
\providecommand{\BIBentrySTDinterwordspacing}{\spaceskip=0pt\relax}
\providecommand{\BIBentryALTinterwordstretchfactor}{4}
\providecommand{\BIBentryALTinterwordspacing}{\spaceskip=\fontdimen2\font plus
\BIBentryALTinterwordstretchfactor\fontdimen3\font minus
  \fontdimen4\font\relax}
\providecommand{\BIBforeignlanguage}[2]{{%
\expandafter\ifx\csname l@#1\endcsname\relax
\typeout{** WARNING: IEEEtran.bst: No hyphenation pattern has been}%
\typeout{** loaded for the language `#1'. Using the pattern for}%
\typeout{** the default language instead.}%
\else
\language=\csname l@#1\endcsname
\fi
#2}}
\providecommand{\BIBdecl}{\relax}
\BIBdecl

\bibitem{fragapane2020increasing}
G.~Fragapane, D.~Ivanov, M.~Peron, F.~Sgarbossa, and J.~O. Strandhagen,
  ``Increasing flexibility and productivity in industry 4.0 production networks
  with autonomous mobile robots and smart intralogistics,'' \emph{Annals of
  operations research}, pp. 1--19, 2020.

\bibitem{alatise2020review}
M.~B. Alatise and G.~P. Hancke, ``A review on challenges of autonomous mobile
  robot and sensor fusion methods,'' \emph{IEEE Access}, vol.~8, pp.
  39\,830--39\,846, 2020.

\bibitem{dugas2020navrep}
D.~Dugas, J.~Nieto, R.~Siegwart, and J.~J. Chung, ``Navrep: Unsupervised
  representations for reinforcement learning of robot navigation in dynamic
  human environments,'' \emph{arXiv preprint arXiv:2012.04406}, 2020.

\bibitem{guldenringlearning}
R.~G{\"u}ldenring, M.~G{\"o}rner, N.~Hendrich, N.~J. Jacobsen, and J.~Zhang,
  ``Learning local planners for human-aware navigation in indoor
  environments.''

\bibitem{everett2018motion}
M.~Everett, Y.~F. Chen, and J.~P. How, ``Motion planning among dynamic,
  decision-making agents with deep reinforcement learning,'' in \emph{2018
  IEEE/RSJ International Conference on Intelligent Robots and Systems
  (IROS)}.\hskip 1em plus 0.5em minus 0.4em\relax IEEE, 2018, pp. 3052--3059.

\bibitem{chen2019crowd}
C.~Chen, Y.~Liu, S.~Kreiss, and A.~Alahi, ``Crowd-robot interaction:
  Crowd-aware robot navigation with attention-based deep reinforcement
  learning,'' in \emph{2019 International Conference on Robotics and Automation
  (ICRA)}.\hskip 1em plus 0.5em minus 0.4em\relax IEEE, 2019, pp. 6015--6022.

\bibitem{faust2018prm}
A.~Faust, K.~Oslund, O.~Ramirez, A.~Francis, L.~Tapia, M.~Fiser, and
  J.~Davidson, ``Prm-rl: Long-range robotic navigation tasks by combining
  reinforcement learning and sampling-based planning,'' in \emph{2018 IEEE
  International Conference on Robotics and Automation (ICRA)}.\hskip 1em plus
  0.5em minus 0.4em\relax IEEE, 2018, pp. 5113--5120.

\bibitem{chiang2019learning}
H.-T.~L. Chiang, A.~Faust, M.~Fiser, and A.~Francis, ``Learning navigation
  behaviors end-to-end with autorl,'' \emph{IEEE Robotics and Automation
  Letters}, vol.~4, no.~2, pp. 2007--2014, 2019.

\bibitem{kastner2021towards}
L.~K{\"a}stner, T.~Buiyan, X.~Zhao, L.~Jiao, Z.~Shen, and J.~Lambrecht,
  ``Towards deployment of deep-reinforcement-learning-based obstacle avoidance
  into conventional autonomous navigation systems,'' \emph{arXiv preprint
  arXiv:2104.03616}, 2021.

\bibitem{rosmann2015timed}
C.~R{\"o}smann, F.~Hoffmann, and T.~Bertram, ``Timed-elastic-bands for
  time-optimal point-to-point nonlinear model predictive control,'' in
  \emph{2015 european control conference (ECC)}.\hskip 1em plus 0.5em minus
  0.4em\relax IEEE, 2015, pp. 3352--3357.

\bibitem{rosmann2019time}
C.~R{\"o}smann, ``Time-optimal nonlinear model predictive control,'' Ph.D.
  dissertation, Dissertation, Technische Universit{\"a}t Dortmund, 2019.

\bibitem{fox1997dynamic}
D.~Fox, W.~Burgard, and S.~Thrun, ``The dynamic window approach to collision
  avoidance,'' \emph{IEEE Robotics \& Automation Magazine}, vol.~4, no.~1, pp.
  23--33, 1997.

\bibitem{chen2017decentralized}
Y.~F. Chen, M.~Liu, M.~Everett, and J.~P. How, ``Decentralized
  non-communicating multiagent collision avoidance with deep reinforcement
  learning,'' in \emph{2017 IEEE international conference on robotics and
  automation (ICRA)}.\hskip 1em plus 0.5em minus 0.4em\relax IEEE, 2017, pp.
  285--292.

\bibitem{milutinovic2018markov}
D.~Milutinovi{\'c}, D.~W. Casbeer, and M.~Pachter, ``Markov inequality rule for
  switching among time optimal controllers in a multiple vehicle intercept
  problem,'' \emph{Automatica}, vol.~87, pp. 274--280, 2018.

\bibitem{toibero2011switching}
J.~M. Toibero, F.~Roberti, R.~Carelli, and P.~Fiorini, ``Switching control
  approach for stable navigation of mobile robots in unknown environments,''
  \emph{Robotics and Computer-Integrated Manufacturing}, vol.~27, no.~3, pp.
  558--568, 2011.

\bibitem{jin2017stable}
J.~Jin, Y.~Kim, S.~Wee, D.~Lee, and N.~Gans, ``A stable switched-system
  approach to collision-free wheeled mobile robot navigation,'' \emph{Journal
  of Intelligent \& Robotic Systems}, vol.~86, no. 3-4, pp. 599--616, 2017.

\bibitem{malone2017hybrid}
N.~Malone, H.-T. Chiang, K.~Lesser, M.~Oishi, and L.~Tapia, ``Hybrid dynamic
  moving obstacle avoidance using a stochastic reachable set-based potential
  field,'' \emph{IEEE Transactions on Robotics}, vol.~33, no.~5, pp.
  1124--1138, 2017.

\bibitem{jacobson2021approximate}
R.~Jacobson and G.~Droge, ``Approximate filippov-based switching for
  behavior-based trajectory tracking,'' in \emph{2021 American Control
  Conference (ACC)}.\hskip 1em plus 0.5em minus 0.4em\relax IEEE, 2021, pp.
  834--839.

\bibitem{berkane2019hybrid}
S.~Berkane, A.~Bisoffi, and D.~V. Dimarogonas, ``A hybrid controller for
  obstacle avoidance in an $ n $-dimensional euclidean space,'' in \emph{2019
  18th European Control Conference (ECC)}.\hskip 1em plus 0.5em minus
  0.4em\relax IEEE, 2019, pp. 764--769.

\bibitem{cimurs2020goal}
R.~Cimurs, J.~H. Lee, and I.~H. Suh, ``Goal-oriented obstacle avoidance with
  deep reinforcement learning in continuous action space,'' \emph{Electronics},
  vol.~9, no.~3, p. 411, 2020.

\bibitem{zhang2019behavior}
W.~Zhang and Y.~Zhang, ``Behavior switch for drl-based robot navigation,'' in
  \emph{2019 IEEE 15th International Conference on Control and Automation
  (ICCA)}.\hskip 1em plus 0.5em minus 0.4em\relax IEEE, 2019, pp. 284--288.

\bibitem{kastner2020deep}
L.~K{\"a}stner, C.~Marx, and J.~Lambrecht, ``Deep-reinforcement-learning-based
  semantic navigation of mobile robots in dynamic environments,'' in \emph{2020
  IEEE 16th International Conference on Automation Science and Engineering
  (CASE)}.\hskip 1em plus 0.5em minus 0.4em\relax IEEE, 2020, pp. 1110--1115.

\end{thebibliography}

\end{document}